\documentclass{article}
\usepackage{styles/ijcai25}

\usepackage{times}
\usepackage{soul}
\usepackage{url}
\usepackage[hidelinks]{hyperref}
\usepackage[utf8]{inputenc}
\usepackage[small]{caption}
\usepackage{graphicx}
\usepackage{amsmath}
\usepackage{amsthm}
\usepackage{booktabs}
\usepackage{nicefrac}
\usepackage{svg}
\usepackage[algoruled, linesnumbered, noend]{algorithm2e}
\usepackage[switch]{lineno}
\usepackage{tikz}
\usetikzlibrary{patterns,arrows,arrows.meta,calc,shapes,shadows,decorations.pathmorphing,decorations.pathreplacing,automata,shapes.multipart,positioning,shapes.geometric,fit,circuits,trees,shapes.gates.logic.US,fit}
\usetikzlibrary{backgrounds,scopes}

\newcommand{\mathsymbol}[2]{ \newcommand{#1}{\ensuremath{#2}\xspace} }

\newcommand{\reals}{\mathbb{R}}
\newcommand{\unitinterval}{[0,1]}
\newcommand{\supp}{\mathrm{supp}}

\newcommand{\Distr}[1]{\Delta(#1)}

\newcommand{\tuple}[1]{\langle #1 \rangle}

\mathsymbol{\sinit}{s_{0}}
\mathsymbol{\Act}{A}
\mathsymbol{\act}{a}
\mathsymbol{\mpm}{T}

\mathsymbol{\obsfun}{O}
\mathsymbol{\obs}{z}
\mathsymbol{\obsset}{Z}
\mathsymbol{\reward}{R}

\mathsymbol{\mc}{C}
\mathsymbol{\pomdp}{M}
\mathsymbol{\worstpomdp}{\underline{M}}
\mathsymbol{\hmpomdp}{\mathcal{M}}

\newcommand{\pomdpT}{\tuple{S,\sinit,\Act,\mpm,\reward,\obsset,\obsfun}}
\newcommand{\mdpT}{\tuple{S,\sinit,\Act,\mpm,\reward}}
\newcommand{\mcT}{\tuple{S,\sinit,\mpm,\reward}}
\newcommand{\hmpomdpT}{\tuple{S,\sinit,\Act,\{\mpm_i\}_{i \in \indices},\{\reward_i\}_{i \in \indices},\obsset,\obsfun}}

\mathsymbol{\imc}{\pomdp^\fsc}
\mathsymbol{\ireward}{\reward^\fsc}
\mathsymbol{\impm}{\mpm^\fsc}

\DeclareMathOperator*{\argmax}{argmax}
\DeclareMathOperator*{\argmin}{argmin}

\DeclareMathOperator{\proj}{proj}

\newcommand{\schedrandom}[1][]{\sigma_{\textrm{rand}}}

\mathsymbol{\policies}{\Pi^{\hmpomdp}}

\mathsymbol{\indices}{\mathcal{I}}

\newcommand{\widesim}[1][1.5]{ \mathrel{ \scalebox{#1}[1]{$\sim$} } }
\mathsymbol{\similarto}{ \,\substack{s,\act \\ \widesim[1.5] }\,}

\mathsymbol{\indicespartition}{\indices /\! \similarto}
\mathsymbol{\subindices}{I}

\mathsymbol{\family}{\mathcal{F}^{\hmpomdp}}
\renewcommand{\path}{\xi}

\mathsymbol{\pmp}{\mathcal{P}}

\mathsymbol{\qpomdp}{ \mathcal{Q} }
\mathsymbol{\qpomdpT}{ \tuple{S,\sinit,\qAct,\qmpm, \qreward, \obsset, \obsfun} }
\mathsymbol{\qAct}{ \Act^{\qpomdp} }
\mathsymbol{\qmpm}{ \mpm^{\qpomdp} }
\mathsymbol{\qreward}{ \reward^{\qpomdp} }

\mathsymbol{\actnought}{ \varnothing }
\mathsymbol{\Actnought}{ \Act_{\actnought} }
\mathsymbol{\lmdp}{ \mathcal{L} }
\mathsymbol{\lAct}{ \Act^\lmdp }
\mathsymbol{\lmpm}{ \mpm^\lmdp }
\mathsymbol{\lreward}{ \reward^\lmdp }

\mathsymbol{\coloring}{ \Gamma }

\mathsymbol{\ptree}{\mathcal{T}}
\mathsymbol{\nodelabel}{F}
\mathsymbol{\leaflabel}{L}

\mathsymbol{\game}{\mathcal{G}}

\mathsymbol{\reachable}{\mathsf{Reach}}
\mathsymbol{\consdindices}{\mathsf{ConsIDs}}

\mathsymbol{\indicesone}{\indices_1}
\mathsymbol{\indicestwo}{\indices_2}

\mathsymbol{\familyone}{\family_1}
\mathsymbol{\familytwo}{\family_2}

\newcommand{\robustobjective}{\mathcal{J}^{\fsc}_{\hmpomdp}}
\newcommand{\pomdpobjective}{J^{\fsc}_{\pomdp}}
\newcommand{\worstpomdpobjective}{J^{\fsc}_{\worstpomdp}}
\newcommand{\pomdpobjectivei}{J^{\fsc}_{\pomdp_i}}

\newcommand{\enumsuffix}{-E}
\newcommand{\unionsuffix}{-U}
\newcommand{\tool}[1]{{\textsc{#1}}}

\newcommand{\paynt}{\tool{Paynt}\xspace}
\newcommand{\ours}{\tool{rfPG}\xspace}
\newcommand{\oursS}{\tool{rfPG-S}\xspace}
\newcommand{\saynt}{\tool{Saynt}\xspace}
\newcommand{\gd}{\tool{ga}\xspace}
\newcommand{\sayntE}{\tool{Saynt\enumsuffix}\xspace}
\newcommand{\gdE}{\tool{ga\enumsuffix}\xspace}
\newcommand{\sayntU}{\tool{Saynt\unionsuffix}\xspace}
\newcommand{\gdU}{\tool{ga\unionsuffix}\xspace}

\newcommand{\gauss}{\mathcal{N}}
\DeclareMathOperator{\clip}{clip}

\def\orcidID#1{\smash{\href{http://orcid.org/#1}{\protect\raisebox{-1.25pt}{\protect\includegraphics{figures/orcid_color.eps}}}}}

\newcommand{\maris}[1]{\textcolor{olive}{MG: #1}}
\newcommand{\iter}{k}

\mathsymbol{\fsc}{\pi}
\mathsymbol{\fscT}{ \tuple{N, n_0, \delta, \eta }}
\newcommand{\fscparamT}[1]{ \tuple{N, n_0, \delta_{\theta^{#1}}, \eta_{\phi^{#1}}}}

\mathsymbol{\gdsteps}{ \textsc{gaSteps} }

\renewcommand{\paragraph}[1]{\smallskip\noindent\emph{#1}}
\renewcommand{\subsubsection}[1]{\medskip\noindent\textbf{#1}}

\newcounter{it}
\def\mylines{}%
\loop\ifnum\theit<4
  \addtocounter{it}{1}
  \expandafter\def\expandafter\mylines\expandafter{%
    \mylines
    \cmidrule(lr){\value{it}-\value{it}}
  }%
\repeat

\SetCommentSty{mycommfont}

\usepackage{adjustbox}
\usepackage{calc}
\newlength\myheight
\newlength\mydepth
\settototalheight{\myheight}{Xygp}
\newif\ifappendix
\global\appendixtrue

\usepackage{cleveref}
\usepackage{xcolor}
\usepackage{microtype}
\usepackage{pgfplots}
\usepackage{amsfonts}
\usepackage{amsmath,amssymb,amsfonts}
\usepackage{mathtools}
\usepackage{mdframed}
\usepackage{enumitem}
\usepackage{xspace}
\usepackage{caption}
\usepackage{subcaption}
\usepackage{multirow}

\usepackage{booktabs,siunitx}
\usepackage{comment}

\newtheorem{example}{Example}

\newtheorem{definition}{Definition}
\newtheorem{remark}{Remark}

\title{
Robust Finite-Memory Policy Gradients for Hidden-Model POMDPs
}

\author{
Maris~F.~L.~Galesloot$^1$\and%
Roman~Andriushchenko$^2$\and%
Milan~\v{C}e\v{s}ka$^{2}$\and%
Sebastian~Junges$^1$\And%
Nils~Jansen$^{3,1}$\\
\affiliations
$^1$Radboud University Nijmegen, The Netherlands\\
$^2$Brno University of Technology, Czechia\\
$^3$Ruhr-University Bochum, Germany\\
\emails
\{maris.galesloot,\,sebastian.junges\}@ru.nl,
\{iandri,\,ceskam\}@fit.vutbr.cz,
n.jansen@rub.de
}

\begin{document}

\maketitle

\begin{abstract}
Partially observable Markov decision processes (POMDPs) model specific environments in sequential decision-making under 
uncertainty.
Critically, optimal policies for POMDPs may not be robust against perturbations in the environment.
Hidden-model POMDPs (HM-POMDPs) capture sets of different environment models, that is, POMDPs with a shared action and observation space.
The intuition is that the true model is hidden among a set of potential models, and it is unknown which model will be the environment at execution time.
A policy is robust for a given HM-POMDP if it achieves sufficient performance for each of its POMDPs.
We compute such robust policies by combining two orthogonal techniques: (1) a deductive formal verification technique that supports tractable robust policy evaluation by computing a worst-case POMDP within the HM-POMDP, and (2) subgradient ascent to optimize the candidate policy for a worst-case POMDP.
The empirical evaluation shows that, compared to various baselines, our approach (1) produces policies that are more robust and generalize better to unseen POMDPs, and (2) scales to HM-POMDPs that consist of over a hundred thousand environments.



\end{abstract}

\section{Introduction}

Partially observable Markov decision processes (POMDPs) \cite{kaelbling1998planning} are the ubiquitous model in decision-making where agents have to account for 
uncertainty over the current state.
Policies for POMDPs select actions based on observations, which provide limited information about the state, and require memory to act optimally.

\begin{figure*}
    \centering
    \begin{subfigure}[b]{.15\textwidth}
        \centering
        \includegraphics[width=.6\textwidth]{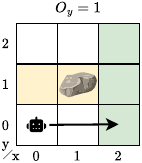}
        \caption{}
        \label{fig:ex:pomdp}
    \end{subfigure}
    \begin{subfigure}[b]{.15\textwidth}
        \centering
        \includegraphics[width=\textwidth]{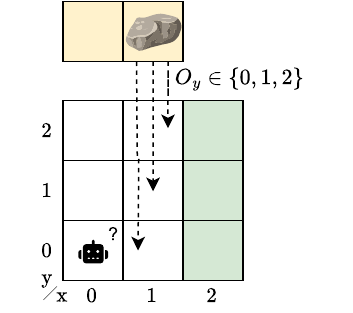}
        \caption{}
        \label{fig:ex:family}
    \end{subfigure}
    \begin{subfigure}[b]{.10\textwidth}
        \centering
        \includegraphics[width=.86\textwidth]{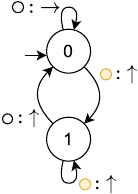}
        \caption{}
        \label{fig:ex:fsc}
    \end{subfigure}
    \begin{subfigure}[b]{.25\textwidth}
        \centering
        \includegraphics[width=0.85\textwidth]{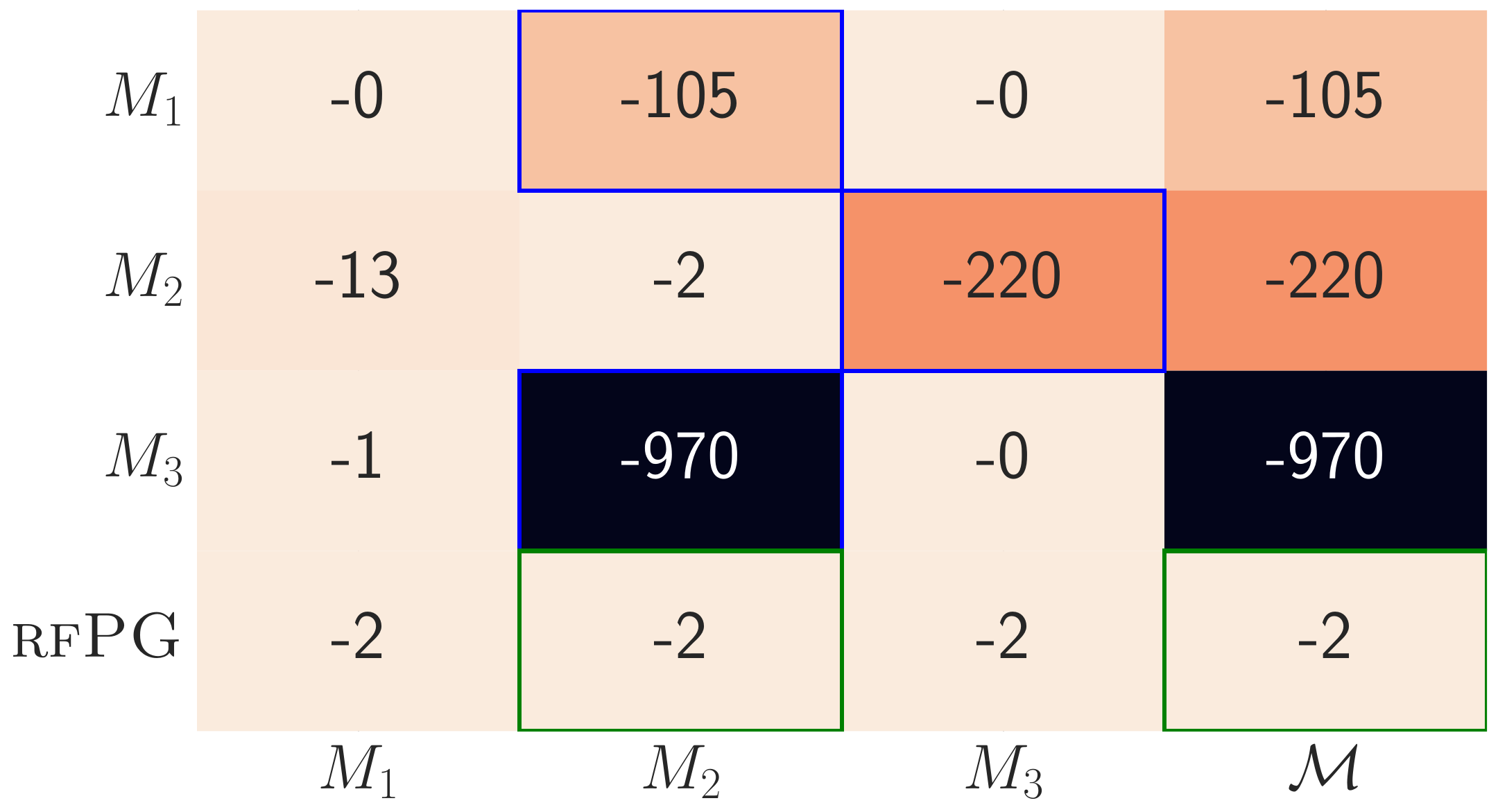}
        \caption{}
        \label{fig:ex:heatmap}
    \end{subfigure}
    \begin{subfigure}[b]{.33\textwidth}
        \centering
        \resizebox{\linewidth}{!}{\newcommand{\nodewidth}{2cm}
\newcommand{\nodegaph}{2.2cm}
\newcommand{\nodegapv}{1cm}
\newcommand{\textfill}[0]{white}
	
	\tikzset{item node/.style={rectangle, draw, text width=2.8cm, text badly centered, node distance=\nodewidth, inner sep=5pt,rounded corners=0.1cm,minimum height=1.5cm,minimum width=2.5cm}}
	\tikzset{policy/.style={diamond,draw,text width=1.0cm,text badly centered,minimum height=0.5cm,rounded corners=0.2cm}}
	\tikzset{data/.style={ellipse,draw,text width=1.0cm,text badly centered,minimum height=1.0cm}}
	\tikzset{joincircle/.style={draw,circle,fill=black,inner sep=0pt, outer sep=0pt,minimum width=0.2cm}}

	\begin{tikzpicture}
		\node[item node, draw = white] (planning) {(1) \textbf{Policy 
        Optimization
        }};
  
		\node[item node, draw = white, very thick, right=\nodegaph of planning] (evaluation) {(2) \textbf{Robust Policy Evaluation}};

		\draw[draw = black, fill=none, rounded corners=0.1cm] (planning.north west) rectangle (planning.south east);
  
		\draw[very thick] (planning.north east) edge[-latex', bend left] node[text width=1.25cm, text centered, above=0.15]{\small FSC $\pi$} (evaluation.north west);

		\draw[very thick] (evaluation.south west) edge[-latex', bend left] node[text width=1.5cm, text centered, below=0.15]{\small \mbox{POMDP $\pomdp$}} (planning.south east);

  		\draw[draw = black, fill=none, rounded corners=0.1cm] (evaluation.north west) rectangle (evaluation.south east);

        \draw[draw = black, dashed, fill=none, rounded corners=0.1cm, label=RPOMDP] ($(evaluation.north west) + (-0.25,0.25)$) rectangle ($(evaluation.south east) + (0.25,-0.25)$);

        \node[draw = none, fill=none, above=0.25 of evaluation] {HM-POMDP};

	\end{tikzpicture}}
        \caption{}
        \label{fig:rfpg}
    \end{subfigure}
    
     \caption{(a) A single POMDP. Cell colours depict possible observations for the agent: green -- exit, yellow -- obstacle in the current row, and white -- no obstacle. (b) An HM-POMDP with three possible obstacle locations. (c) Robust FSC with two memory nodes, optimized by \ours, that solves any possible configuration of this HM-POMDP.
     (d) Robust evaluation of different policies. (e) High-level overview of \ours.}
    \label{fig:main}
\end{figure*}

\begin{example}
\Cref{fig:ex:pomdp} depicts a POMDP with an agent tasked to reach any of the green cells while avoiding the obstacle in the center.
The agent cannot observe its position but can detect if there is an obstacle in its current row.
Due to strong wind blowing southward, there is a small chance the agent moves south instead of the intended direction. 
The optimal policy for this POMDP is straightforward: to always go right.
\end{example}

\paragraph{Robustness.} 
The common assumption that a single, known POMDP model sufficiently captures a system is often unrealistic. 
Furthermore, the optimal policy for this POMDP may not be optimal or perform well on a slightly perturbed model. 
Therefore, it may be beneficial to assume that the system's actual model is only known to be within a set of model variations that are critically different but share certain similarities.

\begin{example}
\Cref{fig:ex:family} depicts a set of POMDPs comprising three potential obstacle locations.
An optimal policy for any of the individual models overfits to that particular obstacle location and fails to solve the task in all three environments.
\end{example}

\paragraph{Hidden-model POMDPs.}
We introduce \emph{hidden-model POMDPs}~(HM-POMDPs) encapsulating multiple different POMDPs.
The ground truth model is assumed to be hidden among the set of POMDPs.
Notably, the POMDPs share the same actions and observations, and therefore, policies are compatible with all of the POMDPs.
The objective is to compute a policy that is \emph{robust} in the sense that it optimizes for the \emph{worst-case} POMDPs within the set.
Consequently, a \emph{robust policy} achieves a lower bound in terms of performance on the set of POMDPs and, therefore, on the ground truth model.

\paragraph{Policy optimization.}
Computing optimal policies for POMDPs requires infinite memory and is undecidable in general~\cite{madani2000undecidability}.
Therefore, we restrict policies to finite memory via finite-state controllers~(FSCs) as policy representations~\cite{DBLP:conf/uai/MeuleauKKC99}.
Computing a robust policy is challenging for HM-POMDPs, as realistic examples may induce large sets of POMDPs, and policies may overfit when optimized for any particular POMDP.
To ensure robustness, the policy must be optimized for the worst-case POMDPs.
Therefore, we seek an approach that generalizes to the whole set of POMDPs by optimizing it on worst-case POMDPs.


\paragraph{Robust policy evaluation.}
A robust policy evaluation is necessary to deduct the worst-case POMDPs from the HM-POMDP and, consequently, provide a lower bound on performance.
The naive approach is to enumerate all POMDPs, but
the set of POMDPs increases rapidly when we encounter many variations in the model, rendering enumeration intractable.
Therefore, a key part of our approach is efficiently performing robust policy evaluation on large, finite sets of~POMDPs.


\subsection{Contributions}
We introduce the \emph{\textbf{robust finite-memory policy gradient}}~(\ours) algorithm.
\Cref{fig:rfpg} provides an overview of the core steps, namely \textbf{policy optimization on worst-case POMDPs} and \textbf{robust policy evaluation}.
In \ours, we represent the policy by an FSC and optimize its parameters through gradient ascent to improve its robust performance.
During each iteration, \ours improves the policy on the worst-case POMDP of the HM-POMDP, akin to subgradient ascent.
The worst-case POMDPs are selected during robust policy evaluation.
We introduce a novel technique that exploits structural similarities between POMDPs to scale to large sets.
We assume that the structural similarity is given by an equivalence relation on the transition and reward functions of the POMDPs and use it to construct a concise representation of the HM-POMDP that enables efficient evaluation via deductive verification.
In an extensive experimental evaluation on both simple and complex HM-POMDPs, we showcase the improvement of \ours over several baselines, both in robust performance and in generalization to unseen models.\footnote{Code is on {Zenodo}~(\url{https://doi.org/10.5281/zenodo.15479642}) and the paper with appendix is on {arXiv}~\cite{hmpomdps}.}
\begin{example}
~\Cref{fig:ex:fsc} illustrates a robust policy found by \ours, with two memory nodes.
It uses memory to deduce the current configuration and solves it close to optimally, moving up at least twice to counter the southward~wind.
\end{example}

\subsection{Related Work}

\paragraph{Models for multiple environments.}
Computing policies for multiple environments has been studied for finite sets of MDPs, known as both \emph{hidden-model MDPs}~(HM-MDPs)~\cite{DBLP:conf/aaai/ChadesCMNSB12,DBLP:conf/amcc/0005ABT19}, \emph{multiple environment MDPs}~(MEMDPs)~\cite{DBLP:conf/fsttcs/RaskinS14,DBLP:conf/aips/ChatterjeeCK0R20}, and \emph{families of MDPs}~\cite{atva-2024}.
We emphasize that, similarly to the different terms for MDPs, our definition of HM-POMDPs could also be considered as \emph{multiple environment POMDPs}~(MEPOMDPs).
Robust POMDPs~(RPOMDPs)~\cite{DBLP:conf/icml/Osogami15} extend robust MDPs~(RMDPs)~\cite{DBLP:journals/mor/Iyengar05,DBLP:journals/mor/WiesemannKR13} and capture a potentially infinite set of POMDPs.
HM-POMDPs form a proper subclass of robust POMDPs, and can approximate RPOMDPs up to finite precision.
Methods for RPOMDPs typically assume (local) convexity and independence over state-action pairs in the set of POMDPs, which results in models where the environment can change completely at each step, which can be overly conservative. 
In contrast, our approach assumes that a worst-case POMDP is picked adversarially at the start and then remains fixed.

\paragraph{Methods for robust policy optimization.}
For RMDPs, various works optimize policies through policy gradients~\cite{DBLP:conf/aaai/Grand-ClementK21,DBLP:conf/icml/LinXD024}, using subgradients~\cite{DBLP:conf/nips/KumarDGLM23,DBLP:conf/l4dc/RickardAM24} or mirror ascent~\cite{DBLP:conf/icml/WangHP23}.
For RPOMDPs, earlier work introduced FSC policy iteration for optimistic (best-case) optimization~\cite{DBLP:journals/soco/NiL13}.
Methods based on value iteration~\cite{DBLP:conf/icml/Osogami15,DBLP:journals/jet/Saghafian18,DBLP:journals/siamjo/NakaoJS21} 
typically do not scale well to large state spaces.
Recent methods for RPOMDPs find robust FSCs through sequential convex programming~\cite{DBLP:conf/aaai/Cubuktepe0JMST21} or by iteratively optimizing a recurrent neural network on worst-case POMDPs~\cite{DBLP:journals/corr/abs-2408-08770}. 
To the best of our knowledge, HM-POMDPs do not yet exist in the literature, and \ours is the first algorithm to tackle robust policy optimization for HM-POMDPs.

\section{Preliminaries}
\label{sec:prelims}
A \emph{distribution} over a countable set $A$ is a~function $\mu \colon A \rightarrow \unitinterval$, s.t.~$\sum_a \mu(a) = 1$ and $\mu(a)\geq0$ for all $a\in A$.
The \emph{support} of $\mu$ is $\supp(\mu) \coloneqq \left\{ a \in A \mid \mu(a) > 0\right\}$
and $a \sim \mu$ denotes $a \in \supp(\mu)$.
The set $\Distr{A}$ contains all distributions over $A$.
$\nabla_{\theta} f$ denotes the gradient of the function $f$ wrt.\ variable $\theta$, and $\proj_A(\cdot)$ denotes the projection onto the set $A$.

\begin{definition}[POMDP]\label{def:pomdp}
A \emph{partially observable Markov decision process~(POMDP)} is a tuple $\pomdp = \pomdpT$ with a finite set $S$ of \emph{states}, an \emph{initial state} $\sinit \in S$, a finite set~$\Act$ of \emph{actions}, a \emph{transition function} $\mpm \colon S \times \Act \rightarrow \Distr{S}$, a \emph{reward function} $\reward \colon S \times \Act \to \mathbb{R}$, a finite set $\obsset$ of \emph{observations} and a deterministic \emph{observation function} $\obsfun \colon S \rightarrow \obsset$\footnote{Observation functions with a distribution over observations can be encoded by deterministic observation functions at the expense of a polynomial increase in the state space~\cite{ChatterjeeCGK16}.}. 

\end{definition}

We will write $\mpm(s' \mid s,\act)$ to denote $\mpm(s,\act)(s')$.
A \emph{Markov decision process}~(MDP) is a POMDP with a unique observation $\obs\in \obsset$ for every $s \in S$.
A \emph{Markov chain}~(MC) is an MDP with $|\Act|=1$.
To simplify notation, MDPs are tuples $\mdpT$ and MCs are tuples $\mcT$. 
A \emph{path} in an MC is a sequence $\path = (s^0, s^1, \ldots)$ of states where $s^0 = \sinit$ and $s^{t+1} \sim \mpm(s^t)$. $\reward(\path) \coloneqq  \sum_{t=0}^{\infty} \reward( s^t )$ denotes the (possibly infinite) cumulative reward for $\path$~\cite{DBLP:books/wi/Puterman94}.



Let $\mc = \mcT$ be an MC.
We consider \emph{reachability reward} objectives: undiscounted infinite-horizon objectives where we accumulate rewards until reaching a set $G \subset S$ of \emph{goal states}~\cite{DBLP:books/wi/Puterman94}.
We assume that every goal state $s_G \in G$ is absorbing and collects no reward: $\mpm(s_G \mid s_G) = 1$ and $\reward(s_G) = 0$; we further assume that every path $\path$ in \mc terminates in $G$.
This definition encompasses infinite-horizon objectives with discounted reward.
Under the assumptions above, $\reward(\path)$ is a well-defined random variable, and its expectation $J_{\mc} \coloneqq \mathbb{E} \left[ \reward(\path) \right]$ will be referred to as the \emph{value of the MC}~\mc.
This value is obtained from the state-values $V_\mc \colon S \to \reals$ by setting $J_{\mc} = V_{\mc}(\sinit)$, after finding the least fixed point of the recursive equation:
$
V_{\mc}(s) = \reward(s) + \sum\nolimits_{{s'\in S}} \mpm (s' \mid s) V_{\mc}(s').
$
%

To represent observation-based policies for POMDPs, we use \emph{finite-state controllers} (FSCs). 
Various types of FSCs exist in the literature~\cite{DBLP:conf/aaai/AmatoBZ10}. 
In this paper, it is convenient to define FSCs as Mealy machines whose output depends on the current node and the most recent observation.

\begin{definition}[Policy]
\label{def:fsc}
A stochastic policy as represented by a finite-state controller is a tuple $\fsc = \fscT$ where $N$ is a finite set of \emph{memory nodes} with the \emph{initial node} $n_0 \in N$,
$\delta \colon N \times \obsset \to \Distr{\Act}$ is the  \emph{action function}, and $\eta \colon N \times \obsset \to \Distr{N}$ is the  \emph{memory update function}. 
Policy \fsc has \emph{full action support} if $\supp(\delta(n,z)) = \Act$ for every $n \in N$ and $\obs\in\obsset$.
\end{definition}

Given state $s$ with observation $z = O(s)$, the policy \fsc executes action $\act \sim \delta(n,\obs)$ associated with the current node $n$ and the current observation $z$. 
The POMDP evolves to some state $s' \sim \mpm(s, \act)$, and the policy evolves to node $n'\sim\eta(n,z)$.
Imposing a policy \fsc onto POMDP $\pomdp$ yields the \emph{induced} Markov chain $\pomdp^\fsc = (S \times N , \tuple{s_0,n_0}, \impm, \ireward)$, with a transition and reward function, using $z = O(s)$, defined as:
\begin{align*}
\impm( \tuple{s',n'} \mid \tuple{s,n} ) &= \eta(n' \mid n, z) \sum_{\act\in\Act}  \delta(a \mid n,z) \mpm(s'\mid s,\act), \\ 
\ireward( \tuple{s,n} ) &= \sum_{\act\in\Act} \delta(a \mid n,z) \reward(s,a).
\end{align*}

The \emph{value of a policy \fsc} for a POMDP \pomdp is the value of the induced MC: $\pomdpobjective \coloneqq J_{\pomdp^\fsc}$.
In this paper, we assume POMDPs in which the goal states are reachable from any state.
This ensures that $V_{\imc}$ is well-defined for any policy \fsc with full action support and is thus differentiable.

\section{Hidden-Model POMDPs}
\label{sec:familymdp}

We present the main problem statement of the paper.
We consider a \emph{hidden-model POMDP}, which describes an indexed set of POMDPs that share state, action, and observation spaces.
\begin{definition}[HM-POMDP]
Let $\indices$ be a finite set of indices.
A \emph{hidden-model POMDP (HM-POMDP)} is a tuple $\hmpomdp = \hmpomdpT$, where $S, s_0, \Act, \obsset, \obsfun$ are as in \Cref{def:pomdp} and $\{\mpm_i\}_{i\in\indices}$ and $\{\reward_i\}_{i\in\indices}$ are indexed sets of transition and reward functions, respectively.
\end{definition}
Given an index $i\in\indices$, an \emph{instance} of an HM-POMDP $\hmpomdp$ is a POMDP $\pomdp_i = \tuple{S,\sinit,\Act,\mpm_i,\reward_i,\obsset,\obsfun}$.
We assume that each POMDP in the HM-POMDP has an initial state (distribution).
Still, the assumption of the shared initial state $\sinit$ or shared observation function $\obsfun$ is non-restrictive, as it can be lifted by introducing intermediate states at a polynomial increase in computational cost. 
Importantly, the POMDPs described by \hmpomdp differ in their transition functions and may thus differ in their topology, i.e., reachable states and observations.
Instances in $\hmpomdp$ have the same set of policies, denoted by \policies.

\begin{definition}[Robust policy performance and optimal policy]
\label{def:robust}
Let \hmpomdp be an HM-POMDP.
The \emph{robust performance} $\robustobjective$ of a policy \fsc is defined as the value of the worst instance and is maximized by an \emph{optimal robust policy} $\fsc^{*}$, defined as:
    \[
    \fsc^{*} \in \argmax_{\fsc\in\policies} \robustobjective,\quad \text{where} \quad \robustobjective \coloneqq \min_{i \in \indices} \pomdpobjectivei.
    \]
\end{definition}



\noindent Then, the key problem tackled in this paper is:
\begin{mdframed}
\textbf{Goal:} Given an HM-POMDP $\hmpomdp$,
find a policy $\fsc^*$ optimizing the robust performance $\robustobjective$.
\end{mdframed}
Our presentation focuses on the worst-case optimization of reachability rewards, i.e., $\argmax\min$ (or $\argmin\max$ when minimizing costs).
The best-case policy performance and its associated policy are defined analogously ($\argmax\max$ or $\argmin\min$), and our method extends to that setting.

The undecidability of the decision variant of this problem follows straightforwardly from the undecidability of infinite-horizon planning for POMDPs~\cite{madani2000undecidability}.
Therefore, we focus on a sound algorithm that aims to find a policy achieving a high robust performance within a reasonable time.

\begin{example}
\label{ex:eval}
We demonstrate robust policy evaluation on the example presented in~\Cref{fig:ex:family}.
We encode the objective (reaching any of the green cells while avoiding the obstacle) as the maximization of reachability reward, where visiting a cell with the obstacle is penalized by the reward of~$-100$.
The table in \Cref{fig:ex:heatmap} reports the expected reward achieved by four policies (rows) on each of the three POMDPs (first three columns); the last column reports the worst-case reward across all POMDPs, i.e., the robust performance.
The first three rows correspond to 2-FSCs, each optimizing the performance in an individual POMDP. Naturally, policy $\fsc_i$ performs well on POMDP $\pomdp_i$. However, it hits the obstacle, on average, at least once for at least one other POMDP in the HM-POMDP.
The last row reports the values achieved by the policy produced by \ours, our method that takes into account many (in this case, all) POMDPs in the HM-POMDP. The robust performance of this policy can be interpreted as hitting the obstacle only once with a probability of at most 2\% across all POMDPs in the~HM-POMDP.
\end{example}

\section{Robust Finite-Memory Policy Gradients}\label{sec:rfpg}
This section presents the \textbf{\emph{robust finite-memory policy gradient}}~(\ours) algorithm to compute robust policies for HM-POMDPs.
We divide the presentation into the following parts.
In \Cref{sec:rfpgmain}, we explain the steps of the main loop (recall \Cref{fig:rfpg}) of \ours.
In \Cref{sec:gd:on:pomdp}, we present the main step in policy optimization, and \Cref{sec:eval} explains contributions towards robust policy evaluation on HM-POMDPs with many instances. In \Cref{sec:rfpgdetails}, we provide additional details.
\subsection{Overview of \ours}\label{sec:rfpgmain}
\ours alternates between the following two main steps:
\begin{itemize}[topsep=2pt,itemsep=2pt]
    \item \emph{Policy optimization}, through policy (sub)gradients, and,
    \item \emph{Robust policy evaluation}, through deductive verification.
\end{itemize}

During \emph{robust policy evaluation}, we select a POMDP whose value coincides with the robust performance $\robustobjective$.
\begin{definition}[Worst-case POMDP]\label{def:worst:pomdp}
    Given a policy $\fsc$, a \emph{worst-case POMDP}~$\worstpomdp$ is an instance of the HM-POMDP \hmpomdp that is a minimizer of \fsc's robust performance $\robustobjective$:
    \[
    \worstpomdp \in \argmin_{\pomdp_i, i \in \indices} J^{\fsc}_{\pomdp_i}, \quad\text{such that}\quad J^{\fsc}_{\worstpomdp} = \robustobjective
    \]
\end{definition}
Given a policy \fsc, computing its robust performance $\robustobjective$ and a worst-case POMDP $\worstpomdp$ analytically is mathematically tractable since HM-POMDP $\hmpomdp$ has finitely many instances.

During \emph{policy optimization}, \ours optimizes the candidate policy \fsc through policy gradient ascent on POMDP \worstpomdp, locally optimizing the candidate policy \fsc for its robust performance. 
As we represent $\pi$ by an FSC, we parameterize  the action function $\delta_{\theta}$ by $\theta \in \Theta \subseteq \Distr{A}^{Z\times N}$ and the update function $\eta_{\phi}$ by $\phi \in \Phi \subseteq \Distr{N}^{Z\times N}$.
Thus, we simultaneously optimize $\fsc$ to learn what to remember and how to act.

\begin{remark}
We observe that $\robustobjective$ is non-differentiable in general due to the minimization over the finite set.
Thus, it may be infeasible to compute the gradient $\nabla_{\theta, \phi} \robustobjective$ to optimize the candidate policy \fsc for robust performance directly. This challenge is circumvented by our iterative approach.
\end{remark}

\subsection{Policy Optimization for HM-POMDPs}
\label{sec:polopt}
Here, we address the first key component of \ours -- optimization of the candidate policies for robust performance.
Our approach builds on \emph{subgradients}, and we optimize \fsc with a subgradient $\nabla_{\theta, \phi} J^{\fsc}_{\worstpomdp}$, where $\worstpomdp$ is a worst-case POMDP for $\fsc$ as in \Cref{def:worst:pomdp}.
If there is no unique worst-case POMDP given \fsc, we select one of them arbitrarily.
Then, for any $k\in\mathbb{N}$, the projected subgradient ascent of the policy's parameters is:
\begin{align*}
\theta^{(\iter+1)} &{=} \proj_{\Theta}\left(\theta^{(\iter)} + \alpha_\iter \nabla_{\theta^{(\iter)}} J^{\pi^{(\iter)}}_{\worstpomdp}\right) {\Bigm|}_{\worstpomdp {\in}\argmin_{\pomdp_i, i \in \indices} J^{\pi^{(\iter)}}_{\pomdp_i}}\\
\phi^{(\iter+1)} &{=} \proj_{\Phi}\left(\phi^{(\iter)} + \alpha_\iter \nabla_{\phi^{(\iter)}} J^{\fsc^{(\iter)}}_{\worstpomdp}\right) {\Bigm|}_{\worstpomdp {\in}\argmin_{\pomdp_i, i \in \indices} J^{\pi^{(\iter)}}_{\pomdp_i}}
\end{align*}
where $\pi^{(\iter)} = \fscparamT{(\iter)}$, and $\alpha_\iter$ are the (diminishing) step sizes.
By iteratively refining the policy on the worst POMDP via subgradients, we may efficiently learn robust behavior and generalize to other (unseen) POMDPs of the HM-POMDP.
Policy gradient ascent on a single POMDP, i.e., for solving $\argmin_{\fsc\in\policies} \pomdpobjective$, converges to a local optimum~\cite{DBLP:conf/uai/MeuleauKKC99,aberdeen2003}.
The situation is more complex for HM-POMDPs since we are maximizing for the minimum across a set of POMDPs.
In particular, $\nabla J^{\fsc}_{\worstpomdp}$ does not guarantee the ascent of $\robustobjective$, i.e., robust performance at each step may not improve monotonically.
This is in part due to the 
fact that there can exist multiple worst-case POMDP, i.e., there might exist multiple subgradients for a policy $\fsc$.
However, the subgradients still provide a meaningful direction for optimizing the robust performance.
Yet, similar to the case of POMDPs, we may not provide global optimality guarantees.
To combat the non-monotonicity, \ours returns the best robust policy found over all iterations until a time-out is reached.
In the following, we detail how we compute the gradients $\nabla J^{\fsc}_{\worstpomdp}$ for worst-case POMDPs $\worstpomdp$ that we use to optimize the candidate policy \fsc.

\subsubsection{Policy gradients on POMDPs with FSCs.}\label{sec:gd:on:pomdp}
For a worst-case POMDP $\worstpomdp$, computing the gradient of the objective $\nabla_{\phi,\theta} \worstpomdpobjective$ with respect to the policy's parameters $\phi$ and $\theta$ enables us to climb the gradient and improve the policy~\cite{DBLP:conf/uai/MeuleauKKC99}.
To ensure the gradients are well-defined, the partial derivatives:
$
\nicefrac{\partial \eta_{\phi}(n'\mid n, \obs)}{\partial \phi_{n,\obs,n'}} 
$
, and,
$\nicefrac{\partial \delta_{\theta}(a\mid n, \obs)}{\partial \theta_{n,\obs,a}},
$
as well as the ratios:
$\nicefrac{\left|\frac{\partial \eta_{\phi}(n'\mid n, \obs)}{\partial \phi_{n,\obs,n'}}\right|}{\eta_{\phi}(n'\mid n, \obs)}$, and,
$\nicefrac{\left|\frac{\partial \delta_{\theta}(a\mid n, \obs)}{\partial \theta_{n,\obs,a}}\right|}{\delta_{\theta}(a\mid n, \obs)}$,
must be uniformly bounded~\cite{DBLP:conf/icml/AberdeenB02}.

These conditions are satisfied under a \emph{softmax parameterization}.
We have that the parameters can range over the real numbers, i.e., we set $\Phi \subseteq \mathbb{R}^{N\times Z\times N}$ and $\Theta \subseteq \mathbb{R}^{N\times Z\times A}$, making the projections $\proj_{\Theta}$ and $\proj_{\Phi}$ trivial.
The \emph{softmax function} $\sigma$ transforms any finite set of real numbers to a categorical distribution over the set.
Given parameters $\phi \in \Phi$, the probabilities $\eta_{\phi}(n'\mid n, \obs)$ are given, for all $n,\obs,n'\in N\times Z \times N$, based on the exponential function $\exp$:
$
\eta_{\phi}(n'\mid n, \obs) = \sigma_{n'}(\phi_{n,\obs}) = \nicefrac{\exp{(\phi_{n,\obs,n'})}}{\sum_{m} \exp{(\phi_{n,\obs,m})}},
$
The partial derivative of a softmax probability with respect to a particular parameter input is:
\[
\frac{\partial \eta_{\phi}(n'\mid n, \obs)}{\partial \phi_{n,\obs,m}} {=} 
\begin{cases}
    \eta_{\phi}(m\mid n, \obs) (1 - \eta_{\phi}(n'\mid n, \obs)) & m = n'\\
    -\eta_{\phi}(m\mid n, \obs) \eta_{\phi}(n'\mid n, \obs) & m \not= n'
\end{cases}
\]







Recall that the value of the policy $\worstpomdpobjective = V_{\worstpomdp^{\fsc}}(\tuple{s_0, n_0})$ is given by the value of the initial state of the induced MC $\worstpomdp^{\fsc}$.
Then, for $\phi$ and $\eta_\phi$, the expression for the partial derivatives of our objective with respect to the individual parameters is:
\begin{align*}
    \frac{\partial \worstpomdpobjective}{\partial \phi_{n,\obs,n'}} 
    =\sum_m \frac{\partial V_{\worstpomdp^{\fsc}}(\tuple{s_0, n_0})}{\partial \eta_{\phi}(m\mid n, \obs)} \cdot \frac{\partial \eta_{\phi}(m\mid n, \obs)}{\partial \phi_{n,\obs,n'}},
\end{align*}
and similarly for $\theta$ and $\delta_\theta$. 
Then, the gradient $\nabla \worstpomdpobjective$ is comprised of the partial derivatives for the individual parameters:
\begin{align*}
    \nabla \worstpomdpobjective &= \left[\nabla_{\phi} \worstpomdpobjective,  \nabla_{\theta} \worstpomdpobjective \right]\\
    &= 
    \left[\left\{\frac{\partial \worstpomdpobjective}{\partial \phi_{n,\obs,n'}} \bigm\vert \forall n,\obs,n'\right\},  \left\{\frac{\partial \worstpomdpobjective}{\partial \theta_{n,\obs,a}} \mid \forall n,\obs,a\right\}\right].
\end{align*}
By computing $\nabla \worstpomdpobjective$, we optimize the parameters of the policy for the POMDPs $\worstpomdp$ selected during robust policy evaluation.
\subsection{Robust Policy Evaluation}
\label{sec:eval}


The robust performance $\robustobjective$ of a given policy~\fsc (see~Def.~\ref{def:robust}) can be computed by enumerating every POMDP $\pomdp_i, i \in \indices$, applying \fsc to obtain the induced MC $\pomdp_i^\fsc$ and computing its value.
In this section, we develop a methodology that avoids this enumeration and makes our approach scale to HM-POMDPs that describe many instances.
The key to such a methodology is a concise representation of HM-POMDPs, namely a succinct representation of a set of transition $\{\mpm_i\}_{i \in \indices}$ and reward $\{\reward_i\}_{i \in \indices}$ functions.

To compactly describe a set of (structurally similar) transition functions, we merge transitions that are shared between multiple instances.
For instance, in the HM-POMDP from \Cref{fig:ex:family}, in the initial position $(x=0,y=0)$ of the agent, all actions have the same immediate effect regardless of the particular instance.
On the other hand, when executing an action from state $(x=1,y=0)$, the agent receives a penalty of -100 in the instance $(O_y=0)$, but receives no penalty in all other POMDPs.
Formally, let $\similarto$ be the equivalence relation on $\indices$ defined as $i \similarto j$ iff $\mpm_i(s,\act) = \mpm_j(s,\act)$ and $\reward_i(s,\act) = \reward_j(s,\act)$,  i.e., executing action $\act$ in POMDPs $M_i$ and $M_j$ has the same effect and yields the same reward. \indicespartition denotes the corresponding equivalence partitioning of \indices wrt.~$\similarto$.
Then, to compactly describe a set of structurally similar POMDPs, we introduce a \emph{quotient POMDP}, which is an extension of \emph{quotient MDPs} (a mild variation of feature MDPs~\cite{DBLP:journals/fac/ChrszonDKB18}) used in~\cite{atva-2024} to reason about families of MDPs.
Intuitively, the quotient POMDP is a POMDP that can execute action $\act$ in state $s$ from an arbitrary $M_i, i \in \indices$.



\begin{definition}[Quotient POMDP]
\label{def:quotient-pomdp}
Given HM-POMDP $\hmpomdp = \hmpomdpT$, 
the \emph{quotient POMDP} associated with \hmpomdp is a POMDP $\qpomdp_\hmpomdp = \qpomdpT$ with actions $\qAct = \Act \times 2^{\indices}$. Let  $(\act,\subindices)$ be denoted as $\act_\subindices$. We define $\qmpm(s' \mid s,\act_\subindices) = \mpm_{i} (s' \mid s,\act)$ and
 $\qreward(s,\act_\subindices) = \reward_i(s,\act)$ where $i \in \subindices$ (arbitrarily).
\end{definition}

While for POMDPs in ~\Cref{def:pomdp} we assume that every action is available in every state, in our implementation, we assume w.l.o.g.\ that every state $s$ is associated with some set $\Act(s) \subseteq \Act$ of \emph{available actions}, i.e., the transition function $\mpm \colon S \times \Act \nrightarrow \Distr{S}$ is partial.
Then, in the quotient POMDP, we assume that $\qAct(s) = \{  \act_\subindices \ \mid \act \in \Act(s), \ \subindices \in \indicespartition \}$, i.e., in state $s$ an agent chooses to play a specific \emph{variant} of action $\act$ available in any $M_i, i \in \indices$.
This allows us to efficiently encode an HM-POMDP where there are actions $\act\in A$ that coincide in many instances $M_i, i \in \indices$.

We lift the notion of the induced Markov chain (\Cref{sec:prelims}) from POMDPs to quotient POMDPs: The \emph{induced quotient MDP} defines the set of MCs describing behavior of~\fsc in the POMDPs captured by the quotient POMDP. Intuitively, in the current state $(s,n)$ of the induced quotient MDP, using $z = O(s)$, an action $\act \sim \delta(n,z)$ is selected and the system transitions to an intermediate state $(s,n,\act)$, in which a specific variant of action $\act$ is selected from the set \indicespartition; then, the state is updated according to $\qmpm(s,\act_\subindices)$ and the memory is updated according to $\eta(n,z)$.

\begin{definition}[Induced quotient MDP]
\label{def:quotient-mdp}
Given quotient POMDP $\qpomdp_\hmpomdp = \qpomdpT$ and policy $\fsc = \fscT$. Let $\Actnought \coloneqq \Act \sqcup \{\actnought\}$ where \actnought is a fresh action.
The \emph{induced quotient MDP} associated with $\qpomdp_\hmpomdp$ and \fsc is an MDP $\lmdp_\hmpomdp^\fsc = \tuple{S \times N \times \Actnought, (\sinit,n_0,\actnought),\lAct,\lmpm,\lreward}$ where $\lAct = \{\actnought\} \sqcup 2^\indices$ with available actions $\lAct(\cdot,\cdot,\actnought) = \{\actnought\}$ and $\lAct(s,\cdot,\act) = \indicespartition$. The transition and reward functions are defined as follows, where, using $z = O(s)$:
\begin{align*}
 \lmpm( \tuple{s,n,\act} \mid \tuple{s,n,\actnought},\actnought ) &= \delta(\act \mid n,z),\\
 \lmpm( \tuple{s',n',\actnought} \mid \tuple{s,n,\act}, \indices) &= \qmpm(s' \mid s,\act_\subindices) \cdot \eta(n' \mid n, z), \\ 
 \lreward( \tuple{s,n,\act},\subindices ) &= \qreward(s,\act_\subindices).
\end{align*}
\end{definition}
Given such a quotient MDP compactly describing a family $\{\pomdp_i^\fsc\}_{i \in \indices}$ of induced MCs, we can use deductive formal verification techniques implemented in the tool \paynt~\cite{DBLP:conf/cav/AndriushchenkoC21} to efficiently identify an MC with the minimal value of $J^{\fsc}_{\pomdp_i} = \robustobjective$, obtaining the robust performance of \fsc and an associated worst-case POMDP $\worstpomdp$.

\subsection{\ours Algorithm}
\label{sec:rfpgdetails}

\setlength{\textfloatsep}{10pt}
\begin{algorithm}[tbp]
\DontPrintSemicolon
\SetKwInOut{Input}{Input}
\SetKwInOut{Output}{Output}
\SetKwInOut{Global}{Global}
\SetKwRepeat{Do}{do}{while}
\SetKwComment{Comment}{$\triangleright \ $}{}
\Input{An HM-POMDP \hmpomdp, a set $G$ of goal states}
\Output{A policy $\fsc^{*}$ achieving the best $\mathcal{J}^{\fsc^{*}}_{\hmpomdp}$}
\Global{number of \gd steps \gdsteps, step sizes $\alpha_k$}
\BlankLine
$N \gets \textsc{sizeOfFSC}(\hmpomdp)$ \Comment{\ifappendix 
See App.~\ref{app:FSCsize}
\else
See App.~B
\fi
} \label{alg:size}
$\theta^{(0)}, \phi^{(0)} \gets \textsc{init}(\Theta,\Phi,N)$, $k \gets 0$, $\nu^* \gets - \infty$ \label{alg:init}\;
$\fsc^{(0)} \gets \tuple{N, n_0, \delta_{\theta^{(0)}}, \eta_{\phi^{(0)}}}$~\Comment{Build policy\hspace{-1em}} \label{alg:build}
$\qpomdp_\hmpomdp \gets \textsc{quotientPOMDP} (\hmpomdp)$~\Comment{Def.\ref{def:quotient-pomdp}} \label{alg:quotient}
\While{\textsc{timeout is not reached} \label{alg:while}} 
{
$\lmdp_\hmpomdp^{\fsc^{(k)}} \gets \textsc{inducedMDP}\left(\qpomdp_\hmpomdp,\fsc^{(k)}\right) $ \Comment{Def.~\ref{def:quotient-mdp}} \label{alg:induced}
$\worstpomdp^{(k)},\mathcal{J}^{\fsc^{(k)}}_{\hmpomdp} \gets \paynt\big(\lmdp_\hmpomdp^{\fsc^{(k)}},G\big)$ \Comment{Sec.~\ref{sec:eval}} \label{alg:eval}
 \lIf{$\mathcal{J}^{\fsc^{(k)}}_{\hmpomdp} > \nu^*$} {$\fsc^{*} \gets \fsc^{(k)}$, $\nu^* \gets \mathcal{J}^{\fsc^{(k)}}_{\hmpomdp} $} \label{alg:updateopt}
\For{$j \gets k \textbf{\emph{ to }} k+\gdsteps-1 $ 
\label{alg:policyupdate}} 
{
$\theta^{(j+1)} \gets \proj_{\Theta}\big(\theta^{(j)} + \alpha_{j} \nabla_{\theta^{(j)}} J_{\worstpomdp^{(k)}}^{\fsc^{(j)}}\big)$ \;
$\phi^{(j+1)} \gets \proj_{\Phi}\big(\phi^{(j)} + \alpha_{j} \nabla_{\phi^{(j)}} J_{\worstpomdp^{(k)}}^{\fsc^{(j)}}\big)$ \;
$\fsc^{(j+1)} \gets \tuple{N, n_0, \delta_{\theta^{(j+1)}}, \eta_{\phi^{(j+1)}}}$
}
$k\gets k+\gdsteps$ \;
}
\Return{$\fsc^{*}$} \;
\caption{The \ours algorithm}
\label{alg:rfpg}
\label{alg}
\end{algorithm}
\Cref{alg:rfpg} outlines the key steps of \ours.
First, we determine the number of nodes for the policy \fsc.
Determining the optimal size is undecidable~\cite{madani2000undecidability}, therefore, we apply a heuristic that solves a small sample set of POMDPs from $\hmpomdp$ to determine an adequate size (\Cref{alg:size}, see \ifappendix\Cref{app:FSCsize}\else
Appendix~B%
\fi\xspace).
We then initialize the gradient ascent parameters (\Cref{alg:init}, see \ifappendix~\Cref{app:setup}
\else
Appendix~A\fi\xspace) and build the corresponding policy \fsc (\Cref{alg:build}).
Finally, we build the quotient POMDP $\qpomdp_\hmpomdp$ (\Cref{alg:quotient}, see~\Cref{def:quotient-pomdp}) that compactly represents the set of POMDPs in the given HM-POMDP~$\hmpomdp$.


The main loop runs until the given timeout is reached. 
We first construct for the current policy \fsc the induced quotient MDP $\lmdp_\hmpomdp^{\fsc^{(k)}}$ (\Cref{alg:induced}, see~\Cref{def:quotient-mdp}). 
The tool \paynt~\cite{DBLP:conf/cav/AndriushchenkoC21} takes this MDP  $\lmdp_\hmpomdp^{\fsc^{(k)}}$ and the given goal states $G$ to compute a worst-case POMDP $\worstpomdp^{(k)}$ and the robust performance $\mathcal{J}^{\fsc^{(k)}}_{\hmpomdp}$ 
(\Cref{alg:eval}, see \Cref{sec:eval}).
We then update the running optimum (\Cref{alg:updateopt}).
Finally, we run \gdsteps gradient ascent steps to update parameters $\phi$ and $\theta$ of the policy \fsc, now on~POMDP $\worstpomdp^{(k)}$ (\Cref{alg:policyupdate}, see \Cref{sec:gd:on:pomdp}).
\gdsteps is a hyperparameter that should be tuned based on the size of the HM-POMDP: having many instances $|\indices|$ slows down the policy evaluation, while many states $|S|$ slows down the gradient update steps. 
In our experiments, we picked $\gdsteps=10$, such that at most 75\% of the computation time is spent on policy evaluation.

\section{Experimental Evaluation}
\def\mathdefault#1{#1}
\everymath=\expandafter{\the\everymath\displaystyle}
In this section, we evaluate \ours on the following questions.

\begin{enumerate}[label=\textbf{(Q\arabic*)}]
    \setlength{\itemsep}{1pt}
    \setlength{\parskip}{0pt}
    \setlength{\parsep}{0pt}
    \item Does \ours produce policies with higher robust performance compared to several baselines?
    
    \label{q:comparison}
    \item Can \ours generalize to unseen environments?
    \label{q:generalization}
    \item How does the POMDP selection affect performance?
    \label{q:ablation}
\end{enumerate}

\subsection{Experimental Setting}

\paragraph{Benchmarks.}
We extend four POMDP benchmarks~\cite{Littman1997LearningPF,NPZ17,qui99stochastic} and one family of MDPs~\cite{atva-2024} to HM-POMDPs.
These benchmarks together encompass a varied selection of different complexities of HM-POMDPs, i.e., different numbers of POMDPs and sizes thereof, as reported in \Cref{tab:allresults}.
\ifappendix\Cref{app:benchmarks}
\else
Appendix~C\xspace
\fi
provides a detailed description of the benchmarks.

\paragraph{Baselines.}
For \ref{q:comparison}, we compare our approaches (1) to the POMDP solver \saynt~\cite{andriushchenko23search}, which provides competitive performance to other state-of-the-art tools that compute FSCs, and (2) standard FSC policy gradient ascent~(\gd) for POMDPs~\cite{aberdeen2003,DBLP:conf/vmcai/HeckSJMK22}.
We create four different baselines by using the following two strategies to compute robust policies:%
\begin{itemize} [leftmargin=1em]
    \setlength{\itemsep}{1pt}
    \setlength{\parskip}{0pt}
    \setlength{\parsep}{0pt}
    \item \textbf{Enumeration}: \sayntE/\gdE runs \saynt/\gd independently on each of the POMDPs in the HM-POMDP and selects the policy with the best robust performance.
    \item \textbf{Union}: \sayntU/\gdU runs \saynt/\gd on the \emph{union POMDP}, constructed from the disjoint union of the POMDPs described by the HM-POMDP, with a uniform distribution over the initial states of these POMDPs.
\end{itemize}

\paragraph{Set-up.} 
Due to their size, executing the baselines on the complete sets of POMDPs described by our benchmark suite of HM-POMDPs is infeasible.
Moreover, we are interested in assessing the generalizability of our method \ref{q:generalization}.
Therefore, we design the following experiment: (1) Pick a random subset of ten POMDPs from the full HM-POMDP, (2) compute a robust policy for this smaller \textit{sub-HM-POMDP} using the four baselines and \ours (referred to as \oursS),
(3) compare the achieved robust performance of \ours to the baselines on this sub-HM-POMDP~\ref{q:comparison}.
To further study the scalability and generalization of \ours to large HM-POMDPs, we extend the experiment with the following steps:
(4) compute a robust policy for the full HM-POMDP using \ours, and (5) compare the robust performance of the resulting six policies on the \textit{full HM-POMDP} using the policy evaluation method from \Cref{sec:eval}.
From this experiment, we can not only assess the scalability of our approach compared to the baselines but, moreover, the ability to generalize to unseen environments \ref{q:generalization}. 
Additionally, we can see if \ours produces a better robust performance than \oursS, indicating whether it is essential to assess all POMDPs within an HM-POMDP.
All methods have a one-hour timeout to compute a policy; in case of a timeout, we report the robust performance of a uniform random policy.
To report statistically significant results, each experiment was carried out on 10 different subsets obtained using stratified sampling from the full HM-POMDP.
\ifappendix\Cref{sec:infra}
\else
Appendix~D
\fi
provides information on the infrastructure used to run the experiments.

\subsection{Overview of the Experimental Results}
We present three experimental artifacts that report the key results of the experimental evaluation.

\paragraph{The table in~\Cref{fig:heatmap_new}} reports, similar to \Cref{ex:eval}, the values achieved by the policies on one particular subset of the Obstacles(8,5) HM-POMDP. 
The table shows the performance of the baseline methods except for \gdE, whose results are generally worse and are reported in\ifappendix~\Cref{app:expeval}\else~Appendix~E\fi.
The right-most column shows the results for the full HM-POMDP.
It provides details in this particular environment on how \ref{q:comparison} \ours compares to the baselines on the subset of POMDPs and how \ref{q:generalization} \ours generalizes to the full HM-POMDP.

\begin{figure}[t]
    \renewcommand\sffamily{}
    \resizebox{0.99\linewidth}{!}{\input{figures/plots4/obstacles-8-3-main-body.pgf}}
    \caption{Performance (maximizing) of \sayntE (first 10 rows), \sayntU, \gdU, and {\oursS} for a sub-HM-POMDP of Obstacles(8,5).
    In the first 10 columns, corresponding to the evaluation of the individual POMDPs, %
    we highlight the worst-case value of each method in blue and the best worst-case value among these in green.
    Independently, we highlight the highest value on the full HM-POMDP $\hmpomdp$ (last column) in green.
    For comparison, 
    \ours achieves a value of $-206$ on $\hmpomdp$, surpassing the baselines.
    }
    \label{fig:heatmap_new}
\end{figure}

\begin{table*}[tb]
    \centering
    \small
    \renewcommand{\arraystretch}{0.8}
   
\newcommand{\rules}[0]{
\cmidrule(lr){1-1}
\cmidrule(lr){2-5}
\cmidrule(lr){6-9}
\cmidrule(lr){10-14}
}

\setlength{\tabcolsep}{4.5pt}
\begin{tabular}{l||rrrr|rrrr||rrrrr}
\toprule
\multirow[b]{2}{*}{Model} & \multicolumn{4}{l}{Dimensions} &  \multicolumn{4}{l}{Subset of $|\indices|=10$ POMDPs} & \multicolumn{5}{l}{Full HM-POMDP} \\[.4em]
& $|\indices|$ & $|S|$ & $|\obsset|$ & $|\Act|$ &   \sayntE & \sayntU & \gdE & \gdU & \sayntE & \sayntU & \gdE & \gdU & \oursS \\
\rules
Obstacles(10,2) & $81$ & $299$ & $14$ & $8$ & $0.76$ & $0.81$ & $0.92$ & $0.95$ & $0.19$ & $0.71$ & $0.21$ & $0.94$ & $0.99$ \\
Network & $140$ & $4$k & $20$ & $7$ & $1.06$ & $1.05$ & $0.93$ & $0.96$ & $1.04$ & $1.07$ & $0.72$ & $1.02$ & $1.02$ \\
Avoid & $1600$ & $13$k & $11$ & $10$ & $1.05$ & $0.10$ & $0.52$ & $0.98$ & $0.18$ & $0.18$ & $0.08$ & $1.10$ & $1.23$ \\
\midrule
Rover & $6$k & $86$ & $18$ & $6$ & $0.88$ & $0.85$ & $0.88$ & $0.85$ & $0.75$ & $0.80$ & $0.75$ & $0.80$ & $0.83$ \\
Obstacles(8,5) & $12$k & $380$ & $25$ & $10$ & $0.79$ & $0.67$ & $0.71$ & $0.68$ & $0.62$ & $0.72$ & $0.25$ & $0.81$ & $0.84$ \\
DPM & $131$k & $737$ & $3$ & $5$ & $0.80$ & $0.61$ & $0.95$ & $0.61$ & $0.54$ & $0.64$ & $0.46$ & $0.62$ & $0.91$ \\
\bottomrule
\end{tabular}
\setlength{\tabcolsep}{6pt}

    \caption{Values of policies computed by the baselines and \oursS on a subset of POMDPs of size 10, evaluated on both the subset and the full HM-POMDP.
    Values evaluated on the subset are normalized wrt. the value of \oursS;
    values evaluated on the HM-POMDP are normalized wrt. the value of \ours.
    Each method had a total timeout of 1 hour.
    The reported results are averaged over 10 seeds.
    }
    \label{tab:allresults}
\end{table*}

\paragraph{\Cref{tab:allresults}} summarizes the main results, including all algorithms and benchmarks to provide substantial evidence to answer \ref{q:comparison} and \ref{q:generalization}.
The upper part of the table contains less complex HM-POMDPs, describing a modest number of POMDPs, while the lower part includes more complex problems. 
The left part of the table presents the results for the sub-HM-POMDPs, and the values are normalized wrt.~the value obtained using \oursS. 
The right part presents the results for the full HM-POMDP with the values normalized wrt.~the value obtained using \ours. 
We average values over 10 seeds; the extended\xspace\ifappendix\Cref{app:tab:errorsub,app:tab:errorwhl} in \Cref{app:expeval}
\else
Tables~2 and 3 in Appendix~E
\fi\xspace include standard~errors.
Values below 1 indicate how much the baselines lag behind  \oursS (the left part) and by \ours (the right part), respectively.

\paragraph{The learning curves in~\Cref{fig:curves}} show the robust performance over time for the policies computed by \ours and by the variant of the \gd optimization that in every iteration of~\Cref{alg:rfpg} selects a random POMDP, similar to \emph{domain randomization}~\cite{DBLP:conf/iros/TobinFRSZA17}, to answer \ref{q:ablation}. 
We report curves for two selected benchmarks, a simpler and a harder one; learning curves for all benchmarks are reported in\ifappendix~\Cref{app:expeval}.
\else
~Appendix~E.
\fi
We consider the full HM-POMDP and plot the average values over 10 seeds with 95\% confidence~intervals.

\begin{figure}[tbp]

\renewcommand\sffamily{}
\resizebox{0.49\linewidth}{!}{\input{figures/plots4/network.pgf}}
\resizebox{0.49\linewidth}{!}{\input{figures/plots4/dpm.pgf}}
\caption{Learning curves of \ours compared to a baseline configured to randomly select a POMDP from the HM-POMDP at each iteration.
We plot averages over 10 seeds with 95\% confidence intervals.
}
\label{fig:curves}
\end{figure}

\subsection{Analysis of the Experimental Results}
\subsubsection{\ref{q:comparison} Comparison to baselines}

\paragraph{Detailed evaluation via Obstacles(8,5).}
As expected, the table in~\Cref{fig:heatmap_new} shows that the policy optimized for a particular POMDP achieves excellent relative value on this POMDP, but generally performs poorly on other POMDPs. The best robust performance of \sayntE is achieved with the policy optimized for POMDP $\pomdp_7$, yielding a value of -180 for the subset and -284 for the HM-POMDP.
The baselines using the union POMDP provide more robust policies. 
\mbox{\gdU} outperforms \sayntU in both settings, although \sayntE outperforms \gdE. 
While \sayntU achieves a better average, its robust performance is worse than \gdU. 
Being the best baseline on this benchmark, \gdU computes a policy achieving a value of -173 on the subset and -230 for the HM-POMDP.
Our approach significantly improves these values: \oursS finds a policy with a value of -149 on the subset and -220 on the HM-POMDP, and \ours finds a policy with a value of -206 on the HM-POMDP (not reported in the table).

\paragraph{Comparison on the whole benchmark suite.}
\Cref{tab:allresults} shows that our approach generally outperforms the baselines. 
Below, we analyze the left and right parts of the table in more detail.
    
\paragraph{Subsets.}
The results demonstrate the benefits of our approach already for the subsets of POMDPs, especially for the more complex benchmarks. 
In 4 (out of 6) benchmarks, the best policies produced by the baselines are behind the policies found by \ours. 
We also observe that the relative performance of the particular baselines significantly varies among the benchmarks.
This shows that (i) already small subsets require generalization in the policies for particular POMDPs to obtain robust performance and (ii) the union POMDP, which ``averages" the individual POMDPs in the subset, does not necessarily provide an adequate abstraction and provides worse results than the enumeration even when 
it is tractable to solve it.

\paragraph{Full HM-POMDPs.}
We observe similar trends to the previous setting; however, our approach's benefits are even more significant. 
\gdU and \sayntU are sometimes competitive with \ours on the three benchmarks with smaller $\indices$, as it is more probable that useful POMDPs end up in the union.
For the more complex benchmarks with large $\indices$, in most cases, the baselines cannot produce policies with a value better than 80\% of the value produced by \ours. 
For the most complex benchmark, DPM, the gap between \ours and the best baseline (\sayntU) is 36\%.
We also observe that \ours considerably improves \oursS on these benchmarks, which demonstrates the scalability of our approach: \ours can effectively reason about HM-POMDPs describing over $130000$ environments.

\subsubsection{\ref{q:generalization} Generalization}
Recall~\Cref{tab:allresults} to compare \oursS and the baselines. 
These results demonstrate the effectiveness of policies optimized for the same subset in generalizing to unseen POMDPs within the HM-POMDP. 
Except for the Network benchmark, \oursS provides much better generalization to the full HM-POMDP.
The most complex benchmark, DPM, shows a gap of~26\% between \oursS and the best baseline.

\subsubsection{\ref{q:ablation} Ablation with random POMDP selection}
The learning curves in~\Cref{fig:curves} demonstrate the role of robust policy evaluation within our approach. 
For the Network benchmark with a modest number of POMDPs, the random selection of the POMDP has a decent chance of sampling useful POMDPs for policy optimization, which is similar to what we observed for \gdU and \sayntU. 
Therefore, the performance is competitive with \ours,
yet \ours performs slightly better on average.
On the other hand, for DPM, the HM-POMDP describes over a hundred thousand POMDPs, and the learning curve clearly shows that robust policy evaluation is essential for more complex problems with more POMDPs.
Here, the random selection leads to policies that achieve significantly worse values with very high fluctuation of values during the learning process.

\section{Conclusion}
HM-POMDPs encapsulate sets of POMDPs, with the assumption that the true model is hidden within the set.
The problem is to find a robust policy for all POMDPs.
We separated the concerns of robust policy evaluation to pick worst-case POMDPs and perform gradient ascent on the set of POMDPs to optimize a policy. 
The experimental evaluation confirms that our approach has two key benefits compared to the baselines: (1)~the policies achieve better generalization to unseen POMDPs, and (2) it scales to HM-POMDPs with over one hundred thousand POMDPs. 
As such, the paper presents the first approach that can effectively solve HM-POMDPs, an important and practically-relevant subclass of robust POMDPs.


\appendix

\section*{Ethical Statement}
There are no ethical issues.

\section*{Acknowledgments}
This research has been partially funded by the ERC Starting Grant 101077178 (DEUCE), and the project \mbox{VASSAL}: ``Verification and Analysis for Safety and Security of
Applications in Life'' funded by the European Union under Horizon Europe WIDERA Coordination and Support Action/Grant Agreement No. 101160022. 
Additionally, this work has been partially
funded by the NWO VENI Grant ProMiSe (222.147) and the Czech Science Foundation grant \mbox{GA23-06963S}
(VESCAA).

\begin{figure}[h]
    \centering
    \includegraphics[width=0.2\linewidth]{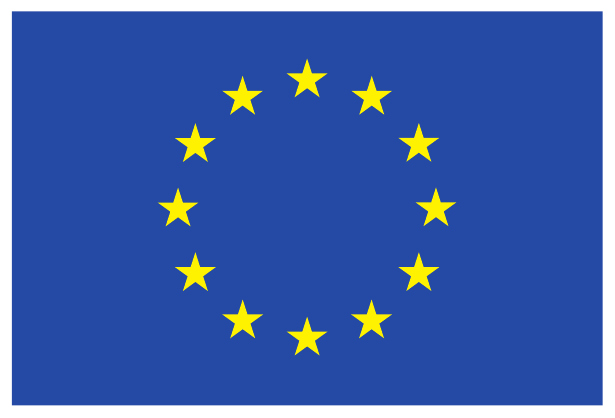}
    \includegraphics[width=0.33\linewidth]{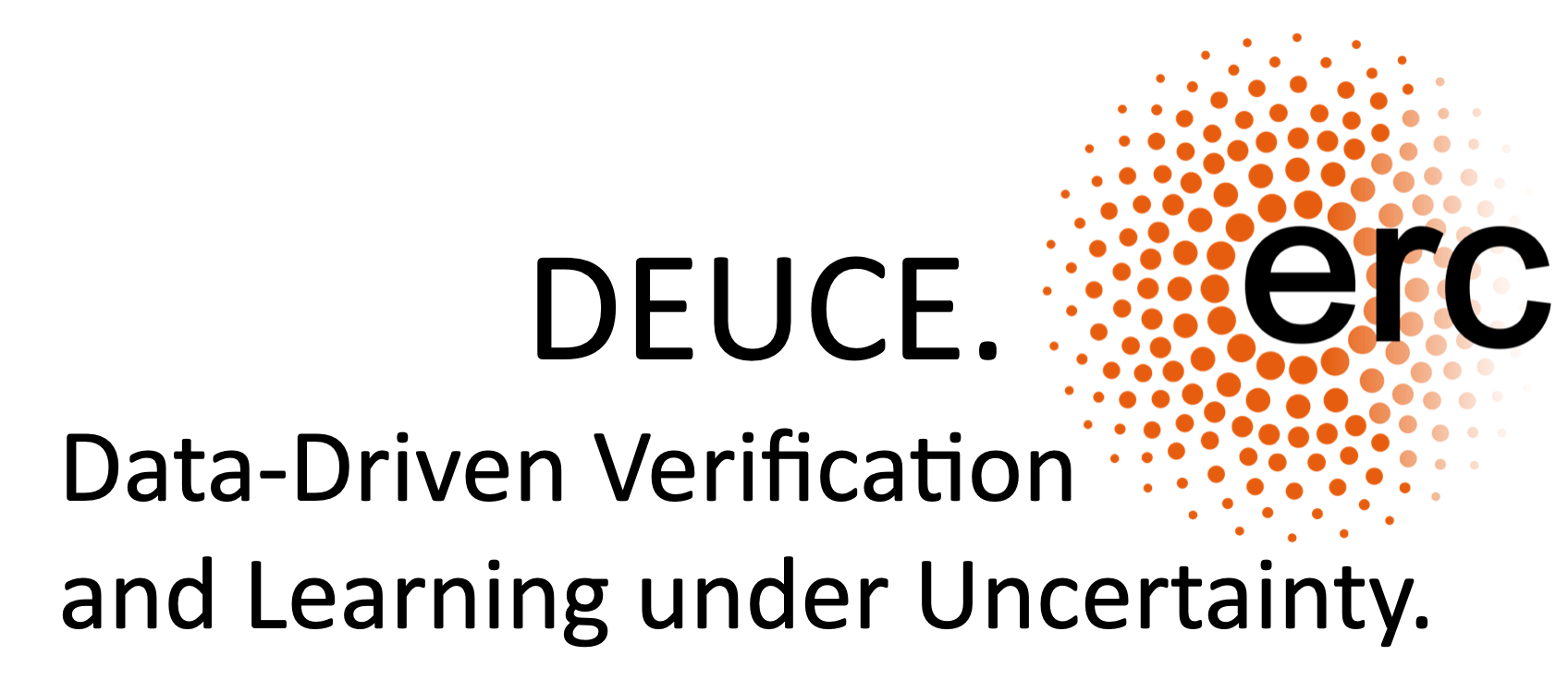}
    \includegraphics[width=0.09\linewidth]{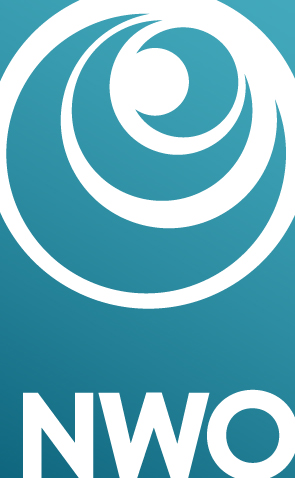}
    \includegraphics[width=0.33\linewidth]{figures/GACR-EN_RGB.png}
    \label{fig:enter-label}
\end{figure}

\bibliographystyle{styles/named.bst}
\bibliography{bibliography}

\begin{thebibliography}{}

\bibitem[\protect\citeauthoryear{Aberdeen and Baxter}{2002}]{DBLP:conf/icml/AberdeenB02}
Douglas Aberdeen and Jonathan Baxter.
\newblock Scalable internal-state policy-gradient methods for {POMDP}s.
\newblock In {\em {ICML}}, pages 3--10. Morgan Kaufmann, 2002.

\bibitem[\protect\citeauthoryear{Aberdeen}{2003}]{aberdeen2003}
Doug Aberdeen.
\newblock {\em Policy-gradient algorithms for partially observable {Markov} decision processes}.
\newblock PhD thesis, The Australian National University, 2003.

\bibitem[\protect\citeauthoryear{Amato \bgroup \em et al.\egroup }{2010}]{DBLP:conf/aaai/AmatoBZ10}
Christopher Amato, Blai Bonet, and Shlomo Zilberstein.
\newblock Finite-state controllers based on {M}ealy machines for centralized and decentralized {{POMDP}s}.
\newblock In {\em {AAAI}}, pages 1052--1058. {AAAI} Press, 2010.

\bibitem[\protect\citeauthoryear{Andriushchenko \bgroup \em et al.\egroup }{2021}]{DBLP:conf/cav/AndriushchenkoC21}
Roman Andriushchenko, Milan {\v{C}}e{\v{s}}ka, Sebastian Junges, Joost-Pieter Katoen, and {\v{S}}imon Stupinsk{\'y}.
\newblock {PAYNT}: a tool for inductive synthesis of probabilistic programs.
\newblock In {\em CAV}, volume 12759 of {\em LNCS}, pages 856--869. Springer, 2021.

\bibitem[\protect\citeauthoryear{Andriushchenko \bgroup \em et al.\egroup }{2023}]{andriushchenko23search}
Roman Andriushchenko, Alexander Bork, Milan {\v{C}}e{\v{s}}ka, Sebastian Junges, Joost-Pieter Katoen, and Filip Mac{\'a}k.
\newblock Search and explore: Symbiotic policy synthesis in {{POMDP}s}.
\newblock In {\em CAV}, 2023.

\bibitem[\protect\citeauthoryear{Andriushchenko \bgroup \em et al.\egroup }{2024}]{atva-2024}
Roman Andriushchenko, Milan \v{C}e{\v{s}}ka, Sebastian Junges, and Filip Mac\'{a}k.
\newblock Policies grow on trees: Model checking families of {MDP}s.
\newblock In {\em ATVA}, 2024.

\bibitem[\protect\citeauthoryear{Chades \bgroup \em et al.\egroup }{2012}]{DBLP:conf/aaai/ChadesCMNSB12}
Iadine Chades, Josie Carwardine, Tara~G. Martin, Samuel Nicol, R{\'{e}}gis Sabbadin, and Olivier Buffet.
\newblock {MOMDPs}: {A} solution for modelling adaptive management problems.
\newblock In {\em {AAAI}}. {AAAI} Press, 2012.

\bibitem[\protect\citeauthoryear{Chatterjee \bgroup \em et al.\egroup }{2016}]{ChatterjeeCGK16}
Krishnendu Chatterjee, Martin Chmel{\'\i}k, Raghav Gupta, and Ayush Kanodia.
\newblock Optimal cost almost-sure reachability in {{POMDP}s}.
\newblock {\em Artif. Intell.}, 234:26--48, 2016.

\bibitem[\protect\citeauthoryear{Chatterjee \bgroup \em et al.\egroup }{2020}]{DBLP:conf/aips/ChatterjeeCK0R20}
Krishnendu Chatterjee, Martin Chmel{\'{\i}}k, Deep Karkhanis, Petr Novotn{\'{y}}, and Am{\'{e}}lie Royer.
\newblock Multiple-environment {Markov} decision processes: Efficient analysis and applications.
\newblock In {\em {ICAPS}}, pages 48--56. {AAAI} Press, 2020.

\bibitem[\protect\citeauthoryear{Chrszon \bgroup \em et al.\egroup }{2018}]{DBLP:journals/fac/ChrszonDKB18}
Philipp Chrszon, Clemens Dubslaff, Sascha Kl{\"{u}}ppelholz, and Christel Baier.
\newblock {ProFeat}: feature-oriented engineering for family-based probabilistic model checking.
\newblock {\em Formal Asp. Comput.}, 30(1):45--75, 2018.

\bibitem[\protect\citeauthoryear{Cubuktepe \bgroup \em et al.\egroup }{2021}]{DBLP:conf/aaai/Cubuktepe0JMST21}
Murat Cubuktepe, Nils Jansen, Sebastian Junges, Ahmadreza Marandi, Marnix Suilen, and Ufuk Topcu.
\newblock Robust finite-state controllers for uncertain {POMDP}s.
\newblock In {\em {AAAI}}, pages 11792--11800. {AAAI} Press, 2021.

\bibitem[\protect\citeauthoryear{Galesloot \bgroup \em et al.\egroup }{2024}]{DBLP:journals/corr/abs-2408-08770}
Maris F.~L. Galesloot, Marnix Suilen, Thiago~D. Sim{\~{a}}o, Steven Carr, Matthijs T.~J. Spaan, Ufuk Topcu, and Nils Jansen.
\newblock Pessimistic iterative planning with {RNN}s for robust {POMDP}s.
\newblock {\em CoRR}, abs/2408.08770, 2024.

\bibitem[\protect\citeauthoryear{Galesloot \bgroup \em et al.\egroup }{2025}]{hmpomdps}
Maris F.~L. Galesloot, Roman Andriushchenko, Milan \v{C}e\v{s}ka, Sebastian Junges, and Nils Jansen.
\newblock Robust finite-memory policy gradients for hidden-model {POMDP}s.
\newblock {\em CoRR}, abs/2505.09518, 2025.

\bibitem[\protect\citeauthoryear{Grand{-}Cl{\'{e}}ment and Kroer}{2021}]{DBLP:conf/aaai/Grand-ClementK21}
Julien Grand{-}Cl{\'{e}}ment and Christian Kroer.
\newblock Scalable first-order methods for robust {MDP}s.
\newblock In {\em {AAAI}}, pages 12086--12094. {AAAI} Press, 2021.

\bibitem[\protect\citeauthoryear{Hartmanns \bgroup \em et al.\egroup }{2020}]{hartmanns20multi}
Arnd Hartmanns, Sebastian Junges, Joost-Pieter Katoen, and Tim Quatmann.
\newblock Multi-cost bounded tradeoff analysis in {MDP}.
\newblock {\em Journal of Automated Reasoning}, 64, 2020.

\bibitem[\protect\citeauthoryear{Heck \bgroup \em et al.\egroup }{2022}]{DBLP:conf/vmcai/HeckSJMK22}
Linus Heck, Jip Spel, Sebastian Junges, Joshua Moerman, and Joost{-}Pieter Katoen.
\newblock Gradient-descent for randomized controllers under partial observability.
\newblock In {\em {VMCAI}}, volume 13182 of {\em Lecture Notes in Computer Science}, pages 127--150. Springer, 2022.

\bibitem[\protect\citeauthoryear{Iyengar}{2005}]{DBLP:journals/mor/Iyengar05}
Garud~N. Iyengar.
\newblock Robust dynamic programming.
\newblock {\em Math. Oper. Res.}, 30(2):257--280, 2005.

\bibitem[\protect\citeauthoryear{Junges \bgroup \em et al.\egroup }{2018}]{DBLP:conf/uai/Junges0WQWK018}
Sebastian Junges, Nils Jansen, Ralf Wimmer, Tim Quatmann, Leonore Winterer, Joost{-}Pieter Katoen, and Bernd Becker.
\newblock Finite-state controllers of {{POMDP}}s using parameter synthesis.
\newblock In {\em {UAI}}, pages 519--529. {AUAI} Press, 2018.

\bibitem[\protect\citeauthoryear{Kaelbling \bgroup \em et al.\egroup }{1998}]{kaelbling1998planning}
Leslie~Pack Kaelbling, Michael~L. Littman, and Anthony~R. Cassandra.
\newblock Planning and acting in partially observable stochastic domains.
\newblock {\em Artificial Intelligence}, 101(1):99--134, 1998.

\bibitem[\protect\citeauthoryear{Kumar \bgroup \em et al.\egroup }{2023}]{DBLP:conf/nips/KumarDGLM23}
Navdeep Kumar, Esther Derman, Matthieu Geist, Kfir~Y. Levy, and Shie Mannor.
\newblock Policy gradient for rectangular robust {Markov} decision processes.
\newblock In {\em NeurIPS}, 2023.

\bibitem[\protect\citeauthoryear{Kwiatkowska \bgroup \em et al.\egroup }{2011}]{KNP11}
Marta Kwiatkowska, Gethin Norman, and David Parker.
\newblock {PRISM} 4.0: Verification of probabilistic real-time systems.
\newblock In {\em CAV}, volume 6806 of {\em LNCS}, pages 585--591. Springer, 2011.

\bibitem[\protect\citeauthoryear{Lin \bgroup \em et al.\egroup }{2024}]{DBLP:conf/icml/LinXD024}
Zhenwei Lin, Chenyu Xue, Qi~Deng, and Yinyu Ye.
\newblock A single-loop robust policy gradient method for robust {Markov} decision processes.
\newblock In {\em {ICML}}. OpenReview.net, 2024.

\bibitem[\protect\citeauthoryear{Littman \bgroup \em et al.\egroup }{1997}]{Littman1997LearningPF}
Michael~L. Littman, Anthony~R. Cassandra, and Leslie~Pack Kaelbling.
\newblock Learning policies for partially observable environments: Scaling up.
\newblock In {\em International Conference on Machine Learning}, 1997.

\bibitem[\protect\citeauthoryear{Madani \bgroup \em et al.\egroup }{2000}]{madani2000undecidability}
Omid Madani, Steve Hanks, and Anne Condon.
\newblock On the undecidability of probabilistic planning and infinite-horizon partially observable {Markov} decision problems.
\newblock {\em Proceedings of the National Conference on Artificial Intelligence}, 02 2000.

\bibitem[\protect\citeauthoryear{Meuleau \bgroup \em et al.\egroup }{1999}]{DBLP:conf/uai/MeuleauKKC99}
Nicolas Meuleau, Kee{-}Eung Kim, Leslie~Pack Kaelbling, and Anthony~R. Cassandra.
\newblock Solving {POMDP}s by searching the space of finite policies.
\newblock In {\em {UAI}}, pages 417--426. Morgan Kaufmann, 1999.

\bibitem[\protect\citeauthoryear{Nakao \bgroup \em et al.\egroup }{2021}]{DBLP:journals/siamjo/NakaoJS21}
Hideaki Nakao, Ruiwei Jiang, and Siqian Shen.
\newblock Distributionally robust partially observable {Markov} decision process with moment-based ambiguity.
\newblock {\em {SIAM} J. Optim.}, 31(1):461--488, 2021.

\bibitem[\protect\citeauthoryear{Ni and Liu}{2013}]{DBLP:journals/soco/NiL13}
Yaodong Ni and Zhi{-}Qiang Liu.
\newblock {Policy iteration for bounded-parameter {POMDP}s}.
\newblock {\em Soft Comput.}, 17(4):537--548, 2013.

\bibitem[\protect\citeauthoryear{Norman \bgroup \em et al.\egroup }{2017}]{NPZ17}
Gethin Norman, David Parker, and Xueyi Zou.
\newblock Verification and control of partially observable probabilistic systems.
\newblock {\em Real-Time Systems}, 53(3):354--402, 2017.

\bibitem[\protect\citeauthoryear{Osogami}{2015}]{DBLP:conf/icml/Osogami15}
Takayuki Osogami.
\newblock Robust partially observable {M}arkov decision process.
\newblock In {\em {ICML}}, volume~37 of {\em {JMLR} Workshop and Conference Proceedings}, pages 106--115. JMLR.org, 2015.

\bibitem[\protect\citeauthoryear{Puterman}{1994}]{DBLP:books/wi/Puterman94}
Martin~L. Puterman.
\newblock {\em {Markov} Decision Processes: Discrete Stochastic Dynamic Programming}.
\newblock Wiley Series in Probability and Statistics. Wiley, 1994.

\bibitem[\protect\citeauthoryear{Qiu \bgroup \em et al.\egroup }{1999}]{qui99stochastic}
Qinru Qiu, Qing Wu, and Massoud Pedram.
\newblock Stochastic modeling of a power-managed system: construction and optimization.
\newblock In {\em Proceedings of the 1999 International Symposium on Low Power Electronics and Design}, page 194–199. Association for Computing Machinery, 1999.

\bibitem[\protect\citeauthoryear{Raskin and Sankur}{2014}]{DBLP:conf/fsttcs/RaskinS14}
Jean{-}Fran{\c{c}}ois Raskin and Ocan Sankur.
\newblock Multiple-environment {Markov} decision processes.
\newblock In {\em {FSTTCS}}, volume~29, pages 531--543. Schloss Dagstuhl - Leibniz-Zentrum f{\"{u}}r Informatik, 2014.

\bibitem[\protect\citeauthoryear{Rickard \bgroup \em et al.\egroup }{2024}]{DBLP:conf/l4dc/RickardAM24}
Luke Rickard, Alessandro Abate, and Kostas Margellos.
\newblock Learning robust policies for uncertain parametric {Markov} decision processes.
\newblock In {\em {L4DC}}, volume 242 of {\em Proceedings of Machine Learning Research}, pages 876--889. {PMLR}, 2024.

\bibitem[\protect\citeauthoryear{Saghafian}{2018}]{DBLP:journals/jet/Saghafian18}
Soroush Saghafian.
\newblock Ambiguous partially observable {M}arkov decision processes: Structural results and applications.
\newblock {\em J. Econ. Theory}, 178:1--35, 2018.

\bibitem[\protect\citeauthoryear{Tobin \bgroup \em et al.\egroup }{2017}]{DBLP:conf/iros/TobinFRSZA17}
Josh Tobin, Rachel Fong, Alex Ray, Jonas Schneider, Wojciech Zaremba, and Pieter Abbeel.
\newblock Domain randomization for transferring deep neural networks from simulation to the real world.
\newblock In {\em {IROS}}, pages 23--30. {IEEE}, 2017.

\bibitem[\protect\citeauthoryear{Wang \bgroup \em et al.\egroup }{2023}]{DBLP:conf/icml/WangHP23}
Qiuhao Wang, Chin~Pang Ho, and Marek Petrik.
\newblock Policy gradient in robust {MDP}s with global convergence guarantee.
\newblock In {\em {ICML}}, volume 202 of {\em Proceedings of Machine Learning Research}, pages 35763--35797. {PMLR}, 2023.

\bibitem[\protect\citeauthoryear{Wiesemann \bgroup \em et al.\egroup }{2013}]{DBLP:journals/mor/WiesemannKR13}
Wolfram Wiesemann, Daniel Kuhn, and Ber{\c{c}} Rustem.
\newblock Robust {M}arkov decision processes.
\newblock {\em Math. Oper. Res.}, 38(1):153--183, 2013.

\bibitem[\protect\citeauthoryear{Wu \bgroup \em et al.\egroup }{2019}]{DBLP:conf/amcc/0005ABT19}
Bo~Wu, Mohamadreza Ahmadi, Suda Bharadwaj, and Ufuk Topcu.
\newblock Cost-bounded active classification using partially observable markov decision processes.
\newblock In {\em {ACC}}, pages 1216--1223. {IEEE}, 2019.

\bibitem[\protect\citeauthoryear{Yang \bgroup \em et al.\egroup }{2011}]{DBLP:conf/globecom/YangMZ11}
Lei Yang, Sugumar Murugesan, and Junshan Zhang.
\newblock Real-time scheduling over {M}arkovian channels: When partial observability meets hard deadlines.
\newblock In {\em GLOBECOM}, pages 1--5, 2011.

\end{thebibliography}

\clearpage
\onecolumn

\ifappendix

\section{Gradient ascent set-up}
\label{app:setup}

We initialize all the parameters in $\phi$ and $\theta$ independently from a zero-mean unit variance Gaussian distribution $\gauss(0, 1)$.
We use a fixed step size, i.e., $\alpha_k = \alpha > 0$, that we set to $\alpha = 0.1$.
We accelerate the ascent using momentum, which considers weighted aggregation of previous derivatives with a decay factor of $\beta=0.9$. 
We do not reset momentum when the worst-case POMDP changes during the iterations.
Furthermore, we clip gradients to a value of $c=5$ to prevent exploding gradients and improve stability during learning.
The momentum is computed as follows, independently for all parameters in $\theta$ and $\phi$.
\begin{align*}
\nu_\theta^{(\iter+1)} &\gets \beta \cdot \nu_\theta^{(\iter)} + (1 - \beta) \cdot \clip(\nabla_\theta \worstpomdpobjective, -c, c),\\
\theta^{(\iter+1)} &\gets \theta^{(\iter)} + \nu_\theta^{(\iter+1)},
\end{align*}
and, equivalently for $\phi$,
\begin{align*}
\nu_\phi^{(\iter+1)} &\gets \beta \cdot \nu_\phi^{(\iter)} + (1 - \beta) \cdot \clip(\nabla_\phi \worstpomdpobjective, -c, c),\\
\phi^{(\iter+1)} &\gets \phi^{(\iter)} + \nu_\phi^{(\iter+1)},
\end{align*}
where we again have $\pi^{(\iter)} = \fscparamT{(\iter)}$.
We compute the gradients by unrolling the FSC onto the POMDP, constructing the resulting \emph{parametric Markov chain}~\cite{DBLP:conf/uai/Junges0WQWK018} and solving a linear equation system to find the partial derivatives~\cite{DBLP:conf/vmcai/HeckSJMK22} that comprise the gradient $\nabla \worstpomdpobjective$.

\section{Determining the controller size}
\label{app:FSCsize}
Since we compute exact gradients for the individual POMDPs, large sizes of FSCs require substantial computation.
Therefore, we opt to optimize sparse FSCs that do not use all memory nodes for each observation.
For a policy $\pi = \fscT$ represented by an FSC, we assume the size is given by a \emph{memory model} $\mu \colon N \times \obsset \to \mathbb{N}$, where $\mu(n,z)$ determines the size of the set $|N|$ for memory node $n\in N$ and observation $\obs\in\obsset$.
We find $\mu$ by running the existing POMDP solver \saynt~\cite{andriushchenko23search} with a short timeout, which can return a candidate memory model from a given POMDP $\pomdp$.
Since we optimize for sets of POMDPs, we sample a \emph{subfamily} of POMDPs from the HM-POMDP.
On each of the $k$ sampled POMDPs, we run \saynt to find the set of memory models $B = \{\hat{\mu}_1, \hat{\mu}_2, \ldots \hat{\mu}_k\}$, and aggregate these results into a single memory model $\mu$ by taking the point-wise maximum $\mu(n, z) = \max_{\hat{\mu} \in B} \hat{\mu}(n,z)$, for all $n,z \in N \times Z$.

We use a stratified sampling scheme without replacement, using the factored representation of variations of the HM-POMDP.
While it does not guarantee that the samples provide practically different POMDPs, it may better cover the differences in the HM-POMDP than a naive uniform sampling approach. 

\section{Benchmark descriptions}
\label{app:benchmarks}

The benchmarks are described using an extension of the guarded-command language PRISM~\cite{KNP11}, where each HM-POMDP is described as a POMDP that contains several parameters (so-called \emph{holes}), similar to~\cite{atva-2024}.
Unlike standard parametric POMDPs, where parameters only affect transition probabilities, holes can also occur in the guard, state update, and rewards associated with the given command.

Obstacles($N$,$X$) and Avoid consider grid worlds~\cite{Littman1997LearningPF} where an agent is tasked with reaching a target cell, similar to the one depicted in~\Cref{fig:ex:family}. Obstacles($N$,$X$) considers an $N$ by $N$ grid with $X$ static obstacles, where holes describe possible obstacle locations.
Similarly, Avoid describes a $5$ by $5$ grid where the agent must reach the target while avoiding adversaries patrolling the grid; the holes determine the initial locations of adversaries as well as their overall behavior.

Network~\cite{NPZ17,DBLP:conf/globecom/YangMZ11} concerns the downlink scheduling of traffic to a number of different users. The model describes a time-slotted system: each packet corresponds to one time period, and these periods are divided into an equal number of slots. The goal is to maximize the number of packets delivered to the users across all the channels. The holes in this HM-POMDP determine the total number of packets/periods and the number of slots in each period.

Rover~\cite{hartmanns20multi,atva-2024} considers scheduling for the Mars rover that has a limited battery and a variety of tasks it can undertake. The tasks differ in their success probability, energy consumption, and scientific value upon success. The goal is to schedule the tasks in order to maximize the scientific value before running out of battery. The holes in the HM-POMDP determine the battery capacity and, for each of the available tasks, its probability of success and energy consumption. This HM-POMDP augments the MDP sketch from~\cite{atva-2024} with limited observability of the rover's battery.

Finally, DPM~\cite{qui99stochastic} describes an electronic component designed to serve incoming requests that arrive stochastically. The component is equipped with a queue request and has multiple power modes (active/idle/asleep), each having different power consumption. The goal is to maximize the number of served requests before running out of battery. The holes in the HM-POMDP determine the speed of switching between modes, the partial observability of the request queue, and the stochastic nature of incoming traffic.

\section{Infrastructure}\label{sec:infra}
All algorithm variants are implemented in the same prototype, implemented in a combination of \texttt{Python3} and \texttt{C++}, and are available in the supplemental material on {Zenodo}~(\url{https://doi.org/10.5281/zenodo.15479642}).
All code ran on the same Linux machine running Ubuntu 22.04.5 LTS, which has an Intel(R) Core(TM) i9-10980XE CPU @ 3.00GHz and 256GB RAM~(8 x 32GB DDR4-3200).
The experiments were executed inside a Docker container running an Ubuntu 22.04 environment.

\section{Additional tables and figures}\label{app:expeval}
This section includes additional results and figures from the experimental evaluation.

\begin{figure}[htbp]
    \begin{center}
    \renewcommand\sffamily{}
    \resizebox{0.9\linewidth}{!}{\input{figures/plots4/obstacles-8-3-GD.pgf}}
    \end{center}
    \caption{Table comparing the performance of the policies computed by the \gd baseline, as well as \oursS, from a single run on Obstacles(8, 5).
    In the first 10 columns, corresponding to the evaluation of the individual POMDPs, we highlight the worst-case value of each method in blue and the best worst-case value among these in green.
    Independently, we highlight the highest value on the whole family (last column) in green.
    For comparison, \ours achieves a value of $-205.9$ on the whole family, surpassing the baselines.
    }
    \label{fig:heatmap_GD}
\end{figure}

In \Cref{app:tab:errorsub,app:tab:errorwhl}, we show the results of \Cref{tab:allresults} of the main paper with standard error for the subfamily and whole family evaluation, respectively.
\begin{table}[tbp]
    \centering
    \begin{tabular}{lllll}
\toprule
Evaluation & \multicolumn{4}{l}{Subset of $|\indices|=10$ POMDPs} \\
Method & \sayntE & \sayntU & \gdE & \gdU \\
\midrule
Obstacles(10,2) & $0.76 \pm 0.04$ & $0.81 \pm 0.06$ & $0.92 \pm 0.04$ & $0.95 \pm 0.05$ \\
Network & $1.06 \pm 0.01$ & $1.05 \pm 0.01$ & $0.93 \pm 0.01$ & $0.96 \pm 0.01$ \\
Avoid & $1.05 \pm 0.13$ & $0.10 \pm 0.01$ & $0.52 \pm 0.07$ & $0.98 \pm 0.08$ \\
\midrule
Rover & $0.88 \pm 0.04$ & $0.85 \pm 0.04$ & $0.88 \pm 0.04$ & $0.85 \pm 0.04$ \\
Obstacles(8,5) & $0.79 \pm 0.06$ & $0.67 \pm 0.02$ & $0.71 \pm 0.05$ & $0.68 \pm 0.05$ \\
DPM & $0.80 \pm 0.04$ & $0.61 \pm 0.04$ & $0.95 \pm 0.04$ & $0.61 \pm 0.04$ \\
\bottomrule
\end{tabular}

    \caption{Extension of \Cref{tab:allresults} part one, for evaluation on the sampled subsets, including standard error with 1 delta degrees-of-freedom.
    The values are normalized with respect to the values achieved by \oursS.}
    \label{app:tab:errorsub}
\end{table}

\begin{table}[tbp]
    \centering

\begin{tabular}{llllll}
\toprule
Evaluation & \multicolumn{5}{l}{Full HM-POMDP} \\
Method & \sayntE & \sayntU & \gdE & \gdU & \oursS \\
\midrule
Obstacles(10, 2) & $0.19 \pm 0.01$ & $0.71 \pm 0.03$ & $0.21 \pm 0.02$ & $0.94 \pm 0.06$ & $0.99 \pm 0.03$ \\
Network & $1.04 \pm 0.01$ & $1.07 \pm 0.01$ & $0.72 \pm 0.02$ & $1.02 \pm 0.01$ & $1.02 \pm 0.01$ \\
Avoid & $0.18 \pm 0.05$ & $0.18 \pm 0.05$ & $0.08 \pm 0.02$ & $1.10 \pm 0.43$ & $1.23 \pm 0.39$ \\
\midrule
Rover & $0.75 \pm 0.00$ & $0.80 \pm 0.00$ & $0.75 \pm 0.00$ & $0.80 \pm 0.00$ & $0.83 \pm 0.02$ \\
Obstacles(8, 5) & $0.62 \pm 0.03$ & $0.72 \pm 0.00$ & $0.25 \pm 0.02$ & $0.81 \pm 0.04$ & $0.84 \pm 0.04$ \\
DPM & $0.54 \pm 0.01$ & $0.64 \pm 0.02$ & $0.46 \pm 0.01$ & $0.62 \pm 0.01$ & $0.91 \pm 0.03$ \\
\bottomrule
\end{tabular}

    \caption{Extension of \Cref{tab:allresults} part two, for evaluation on the whole HM-POMDP, including standard error with 1 delta degrees-of-freedom.
    The values are normalized with respect to the values achieved by \ours.
    }
    \label{app:tab:errorwhl}
\end{table}

\begin{figure*}[tbp]
\renewcommand\sffamily{}
\resizebox{0.32\linewidth}{!}{\input{figures/plots4/obstacles-10-2.pgf}}
\resizebox{0.32\linewidth}{!}{\input{figures/plots4/network.pgf}}
\resizebox{0.32\linewidth}{!}{\input{figures/plots4/avoid.pgf}}
\resizebox{0.32\linewidth}{!}{\input{figures/plots4/rover.pgf}}
\resizebox{0.32\linewidth}{!}{\input{figures/plots4/obstacles-8-3.pgf}}
\resizebox{0.32\linewidth}{!}{\input{figures/plots4/dpm.pgf}}

\caption{Learning curves of \ours compared to a baseline configured to randomly select a POMDP from the HM-POMDP during ascent.
We plot averages over 10 seeds in combination with 95\% confidence intervals.
We negate rewards in environments where the objective is to minimize costs for visibility purposes in the plot.
The upper part of the figure contains the less complex models, while the lower part contains the more complex models. The wide confidence interval at the beginning of \ours in Avoid is caused by a bad initial policy (selected randomly in every seed) that takes a long time to evaluate and thus to improve.
}
\label{fig:curves-full}
\end{figure*}

\fi

\end{document}